\documentclass{article}
\usepackage{spconf,amsmath,graphicx,multirow,hyperref,caption}
\usepackage[absolute,showboxes]{textpos}

\AtBeginDocument{
	\renewcommand{\tablename}{Table}
	\renewcommand{\figurename}{Fig.}
	\newcommand{\figurenarrow}{3.39in}
	
}
\renewenvironment{thebibliography}[1]{
	\begin{oldthebibliography}{#1}
		\setlength{\itemsep}{0.01em}
		\setlength{\parskip}{0.01em}
	}
	{
	\end{oldthebibliography}
}

\TPMargin{5pt}

\newcommand{\copyrightstatement}{
	\begin{textblock}{0.828}(0.088,0.93)    % tweak here: {box width}(leftposition, rightposition)
		\noindent
		\footnotesize
		\textcopyright 2021 IEEE. Personal use of this material is permitted. Permission from IEEE must be obtained for all other uses, in any current or future media, including reprinting/republishing this material for advertising or promotional purposes, creating new collective works, for resale or redistribution to servers or lists, or reuse of any copyrighted component of this work in other works.
	\end{textblock}
}

\begin{document}
\copyrightstatement

\title{Supervised Training of Siamese Spiking Neural Networks\\with Earth Mover's Distance}

\name{Mateusz~Pabian, Dominik~Rzepka, Mirosław~Pawlak\thanks{This work was supported by the Polish National Center of Science under Grant DEC-2017/27/B/ST7/03082.}}
\address{Department of Measurement and Electronics\\
AGH University of Science and Technology, Kraków}

\maketitle

\begin{abstract}
	This study adapts the highly-versatile siamese neural network model to the event data domain. We introduce a supervised training framework for optimizing Earth Mover's Distance (EMD) between spike trains with spiking neural networks (SNN). We train this model on images of the MNIST dataset converted into spiking domain with novel conversion schemes. The quality of the siamese embeddings of input images was evaluated by measuring the classifier performance for different dataset coding types. The models achieved performance similar to existing SNN-based approaches (\mbox{F1-score} of up to 0.9386) while using only about~15\% of hidden layer neurons to classify each example. Furthermore, models which did not employ a sparse neural code were about 45\%~slower than their sparse counterparts. These properties make the model suitable for low energy consumption and low prediction latency applications.

\end{abstract}


\section{Introduction}
\label{sec:01_introduction}

Siamese neural networks are a type of machine learning models which are trained to optimize a similarity measure between network outputs for different examples of data from some domain~\cite{Koch2015}. The resulting model is thus capable of querying inputs based on their similarity, i.e. given some input signal, find the most similar signal or a set of signals. Siamese networks have achieved state-of-the-art results on person re-identification~\cite{Schroff2015}, as well as object-~\cite{Dunnhofer2020} and person-tracking tasks~\cite{Hermans2017}. Furthermore, these networks achieve competitive performance in change point detection~\cite{Bredin2017} and can even be used to design ranking engines~\cite{Liu2017}.

Adapting the siamese model to spiking neural networks (SNN), which process data using asynchronous spikes~\cite{Pfeiffer2018}, necessitates leveraging existing spike train similarity measures developed by the neuroscientific community. Commonly used measures include van~Rossum distance~\cite{Rossum2001}, Schreiber’s similarity measure~\cite{Schreiber2003}, Victor-Purpura distance~\cite{Victor1997} and Earth Mover's Distance (EMD)~\cite{Sihn2019}. Importantly, the choice of a spike train similarity measure is heavily dependent on a chosen neural coding scheme~\cite{Satuvuori2018}. Various schemes have been observed in biological neurons~\cite{Dayan2001}, such as encoding information using impulse frequency~\cite{Adrian1926}, relative timing of the first spike~\cite{Gollisch2008}, or using a group activity pattern across a population of neurons~\cite{Harper2004}. In the context of SNNs the type of neural coding depends on how the network is trained and the chosen neuron model.

Training SNNs such that the output spike train matches a predetermined temporal pattern based on some similarity measure has been studied extensively. Several early studies, such as ReSuMe~\cite{Ponulak2010} or the Tempotron~\cite{Gutig2006}, focused on single-layered networks. More recent works however emphasise the ability to train multilayer SNNs. For example, Lin~\emph{et~al.}~\cite{Lin2017} propose a general supervised learning rule that minimizes an $L_2$-distance between kernel-smoothed spike trains, Zenke~\&~Ganguli~\cite{Zenke2018} optimize the van~Rossum distance between spike trains, whereas Xing~\emph{et~al.}~\cite{Xing2020} train a network by minimizing spike count differences in predetermined time windows over all neurons in the output layer. To the best of our knowledge, the only work directly related to adapting the siamese model to SNNs was summarized by Luo~\emph{et~al.}~\cite{Luo2021}. Their model achieved competitive performance on several visual object tracking benchmarks with low precision loss with respect to the original nonspiking network.

In this paper, we propose a novel supervised training scheme for multilayer siamese SNNs which optimizes EMD between output spike trains~\cite{Sihn2019}. We build the network using neurons tuned to respond to precise timing of input events~\cite{Mostafa2018}. In contrast to the work of Luo~\emph{et~al.}~\cite{Luo2021}, our siamese SNN is optimized in the spiking domain, rather than be a product of converting an existing neural network to the spiking domain. Additionally, we opt to encode information using single events instead of bursts of spiking activity. We train the proposed models on images from the MNIST dataset~\cite{LeCun1998} converted to the spiking domain using a novel input coding scheme based on the concept of implicit information~\cite{Rzepka2018} carried by events in the spiking domain. Finally, we assess the trained models in terms of classification \mbox{F1-score}, hidden layer activation sparsity and \mbox{prediction latency}.

\section{Methods and Algorithms}
\label{sec:02_methods}

\subsection{Backpropagation Algorithm for IF Neural Networks}
\label{sec:021_methods_if}

Let us briefly summarize the model first described in~\cite{Mostafa2018}, which trains a spiking neural network that is sensitive to the timing of input spikes rather than their rate. This type of network uses Integrate-and-Fire (IF) neurons with exponentially decaying synaptic current kernels. For a given kernel-smoothed input spike train signal $x(t)$ with spike-times $\{t_i, i=1,\dots,N\}$
\begin{equation}
	x(t) = \sum_{i=1}^{N} \kappa(t-t_i)
	= \sum_{i=1}^{N} \exp\left(-\frac{t-t_i}{\tau_{syn}}\right)u\left(t-t_i\right),
\end{equation}
where $\tau_{syn}$ is the synaptic current time constant and $u(t)$ denotes a Heaviside step function. We define the IF~neuron dynamics  by the following differential equation
\begin{equation}
\frac{dV(t)}{dt} = \sum_{i=1}^{N} w_i \kappa(t-t_i)
\label{eqn:diff_if_neuron},
\end{equation}
where $V(t)$ is the membrane voltage of the postsynaptic neuron, $i$ is the presynaptic neuron index and $\{w_i\}$ are synaptic weights. Importantly, each presynaptic neuron is associated with exactly one spike-time $t_i$. The closed-form solution of (\ref{eqn:diff_if_neuron}) is given by
\begin{equation}
V(t) = V_0 + \tau_{syn}\sum_{i=1}^{N} w_i \left[1-\exp\left(-\frac{t-t_i}{\tau_{syn}}\right)\right]u\left(t-t_i\right)
\label{eqn:if_neuron},
\end{equation}
where $V_0$ is the initial membrane voltage.

The neuron is said to fire at time $t_{out}$ if the voltage crosses a threshold $V_{thr}$ from below. Then, a subset of input spikes that cause the postsynaptic neuron to fire
\begin{equation}
C = \lbrace i: t_i < t_{out} \rbrace
\label{eq:causal_set}
\end{equation}
is called a \emph{causal set}. Assuming, without the loss of generality,  that $V_0=0$, the solution to~(\ref{eqn:if_neuron}) that describes the relationship between the causal set of input spikes  and the time of a postsynaptic neuron spike $t_{out}$ is given in the implicit form 
\begin{equation}
z_{out} = \frac{\sum_{i \in C}w_i z_i}{\sum_{i \in C}w_i - \frac{V_{thr}}{\tau_{syn}}}
\label{eqn:_if_zdomain_time_to_first_spike},
\end{equation}
where \mbox{$z_i = \exp(\frac{t_i}{\tau_{syn}})$} and $z_{out} = \exp(\frac{t_{out}}{\tau_{syn}})$. This formula is differentiable with respect to synaptic weights $\{w_i\}$ and the transformed input spike times $\{z_i\}$, therefore it can be used to train a spiking network using the backpropagation algorithm.
\subsection{Training a Spiking Siamese Neural Network}
\label{sec:022_methods_siamese}

We train the network to optimize EMD-based similarity between output spike trains~\cite{Sihn2019}. Note that each neuron of the model described in Section \ref{sec:021_methods_if} returns the time of a single event. Therefore, in order to construct the output spike train the events obtained from the last layer are concatenated and sorted in ascending order. We call the resulting point process a \emph{spike train embedding} $f(t)$. EMD is given by
\begin{equation}
\mbox{EMD}(f,g) = \int_{-\infty}^{\infty} \left| F(t) - G(t) \right| dt
\label{eqn:emd}.
\end{equation}
where $F,G$ are cumulative distribution functions of point processes $f,g$, respectively. The function $F$ ($G$ is defined analogously) takes the form $F(t)=\frac{1}{P} \sum_{i=1}^{P} u(t-t_i)$. The distributions $F,G$ are piecewise constant nondecreasing functions, hence the numerical evaluation of~(\ref{eqn:emd}) is straightforward~\cite{Sihn2019}. The overall computational complexity of EMD for a pair of spike trains $f(t)$ and $g(t)$, with $P$ and $R$ events respectively, is $\mathcal{O}(P+R)$, assuming that events $\{ t_{i}, i=1,\dots, P \}$ and $\{  \tau_i, i=1,\dots,R \}$ are sorted \cite{Cohen1999}.

In this work, a triplet-based loss function was used to optimize the network. Let \mbox{$f_a$, $f_p$, $f_n$} be the spike train embeddings of examples \mbox{$a$, $p$, $n$} called the \emph{anchor}, \emph{positive} and \emph{negative} respectively. Then the formula
\begin{equation}
L_{anp} = \max \left( 0, \alpha + \mbox{EMD}(f_a, f_p) - \mbox{EMD}(f_a, f_n) \right)
\label{eq:triplet_loss},
\end{equation}
describes the contribution of a given triplet to the total loss. Minimizing $L_{anp}$ ensures that the distance between $f_a$ and $f_p$ (or the anchor-positive pair of examples) is smaller than the distance between $f_a$ and $f_n$ (anchor-negative pair) by at least some margin~$\alpha$.

Overall, the loss function is a composite of a triplet loss averaged over the set of valid triplets in the mini-batch $Q$ (the `batch-all` strategy from~\cite{Hermans2017}), and the spike regularization term~\cite{Mostafa2018} which promotes network spiking activity by ensuring a nonnegative denominator of~(\ref{eqn:_if_zdomain_time_to_first_spike})
\begin{equation}
\begin{split}
L &= \frac{1}{|Q|} \sum_{q \in Q} L_q + K \sum_{j=1}^{M} \max \left( 0, \frac{V_{thr}}{\tau_{syn}} - \sum_{i=1}^{N} w_{ij} \right)\\
Q &= \lbrace a,n,p :\; a \neq n \neq p ,\; y_a = y_p \neq y_n \rbrace\\
\end{split}
\label{eq:total_loss},
\end{equation}
where \mbox{$y_a$, $y_p$, $y_n$} are the class labels of examples \mbox{$a$, $p$, $n$}, the index $j$ runs over all neurons in the network $M$, and $K$ is a hyperparameter. Finally, let \mbox{$Q_{AT} = \lbrace a,n,p \in Q: L_{anp} \neq 0 \rbrace$} be the set of active triplets which contribute to a nonzero loss for the current batch of examples. Then the ratio of active triplets $\mbox{AT} = \frac{|Q_{AT}|}{|Q|}$ is used as an early stopping criterion for the training procedure.
\section{Results and Discussion}
\label{sec:03_results}

\subsection{Experimental Setup}
\label{sec:030_results_setup}

We test our approach by training siamese SNNs on a MNIST dataset~\cite{LeCun1998} in three seperate settings which differ primarily in the number of events passed down to network input neurons. This allows us to explore how the resulting model properties are influenced by different input data pixel-to-spike conversion methods. Each \mbox{28x28 image} is first flattened into a \mbox{784-element} vector. The resulting vectors are processed differently, depending on the experimental setting:
\\\underline{\emph{Black\&white}} (adapted from~\cite{Mostafa2018}): the vector representation of each image is binarized with a threshold of 50\% of global maximum pixel intensity, and one of two time instants $t_0$ or $t_1$ is assigned to the value of each bit. We set \mbox{$t_0=0$ ms} and \mbox{$t_1=1.79$ ms}, as in~\cite{Mostafa2018}.
\\\underline{\emph{Binary}}: each vector representation is binarized as in the \emph{black\&white} setting, however the `late` event-time $t_1$ is set to infinity (which corresponds to a lack of event for black pixels).		
\\\underline{\emph{Grayscale}}: the original grayscale images are converted to the spiking domain by modeling each pixel as an artificial neuron responding to a driving signal (synaptic current) of constant intensity $I$ proportional to image pixel intensity (in range 0-1). For $V_0=0$, the formula for the membrane voltage of IF-based converter neurons is $V(t)=\frac{t}{\tau_{syn}}I$, and the spike time corresponding to a pixel of a given intensity is
\begin{equation}
	t_{out}=\frac{V_{thr}\tau_{syn}}{I}.
\label{eq:gray_tout}
\end{equation}
This model retains the desirable property that \mbox{$t_{out} \to \infty$} as \mbox{$I \to 0$}.

Note that in \emph{binary} and \emph{grayscale} settings some channels might not have any events associated with them. In this context, a lack of event occurence carries implicit information~\cite{Rzepka2018} that can be exploited by the network.

We use the same \mbox{784-400-400-10} network (denoting the number of neurons in input, hidden and output layers, respectively) in all of our experiments. Each model was optimized using \mbox{RMSprop}~\cite{Tieleman2012} with a learning rate of $10^{-3}$, synapse regularization parameter \mbox{$K=400$}, synaptic time constant \mbox{$\tau_{syn}=1$ ms}, voltage threshold \mbox{$V_{thr}=1$ mV}, and the triplet loss margin \mbox{$\alpha=0.1$}. Additionally, we apply $L_2$ regularization with \mbox{$\lambda=10^{-3}$} to the loss function~(\ref{eq:total_loss}). Finally, a batch size of \mbox{$n=256$} was used across all experiments. No data-augmentation methods were used.

\subsection{MNIST Digit Classification}
\label{sec:031_results_clf}

In order to evaluate our approach, we measure \mbox{k-Nearest} Neighbour \mbox{(k-NN)} classifier performance as a proxy for embedding space example proximity. For a trained model, a spike train embeddings was computed for each example, then $k$ training-set embeddings closest to a given test-set embedding were selected, which then could be used to determine the test example label prediction using majority voting. We find that the classifier performance is not influenced by changing $k$ for $k \geq 7$, therefore we set $k=7$ for all experiments. Compared to other SNN-based approaches (\tablename{~\ref{tab:mnist_summary}}), our best-performing model achieved a similar level of performance using a novel approach to dataset encoding and signal transformation defined by the trained spiking neural network. Importantly, the focus of this work was to show that the proposed methodology can be used to train multilayer spiking siamese neural networks with timing-sensitive neural coding. Obtaining high accuracy was a secondary objective, mainly as a proxy for determining whether the training procedure was successful or not.

\begin{table}[t!]
	\centering
	\caption{Classifier performance of different SNNs on MNIST\newline(best reported results from each paper).}
	\label{tab:mnist_summary}
	\begin{tabular}{c|c|c}
		\multirow{2}{*}{\shortstack[c]{input\\ encoding}}     & \multirow{2}{*}{model type}                       & \multirow{2}{*}{performance}     \\
		& & \\ \hline
		-  & k-NN Euclidean baseline \cite{LeCun1998}         & 0.950\\ \hline
		\multirow{4}{*}{\shortstack[c]{firing\\ rate}} & Spiking RBM \cite{Neftci2014}         & 0.926           \\
		& STPD-trained network \cite{Diehl2015_unsupervised} & 0.950           \\
		& Spiking NN \cite{Diehl2015_fast}      & 0.986           \\
		& \textbf{Spiking CNN} \cite{Diehl2015_fast}        & \textbf{0.991}  \\ \hline
		\multirow{2}{*}{\shortstack[c]{spike-\\ time}}  & Spiking NN \cite{Mostafa2018}         & 0.975           \\
		& Siamese Spiking NN (our)              & 0.946     
	\end{tabular}
\end{table}

\subsection{Hidden Layer Activation Sparsity}
\label{sec:032_results_sparsity}

Interestingly, we have observed that models trained in the \emph{binary} and \emph{grayscale} settings exhibit sparse internal representation of the input signal. We call a group of neurons \emph{quiescent} when they do not fire for a given input signal $x(t)$, or equivalently \mbox{$H_{q,x} = \lbrace h \in H, t>0: V_{h,x}(t) < V_{thr} \rbrace$}. Let the \emph{network sparsity index} ${QN}_x$ be the fraction of quiescent neurons $H_{q,x}$ to all neurons in hidden layers $H$
\begin{equation}
{QN}_x = \frac{|H_{q,x}|}{|H|}.
\end{equation}

\begin{figure}[t!]
	\includegraphics[width=\figurenarrow]{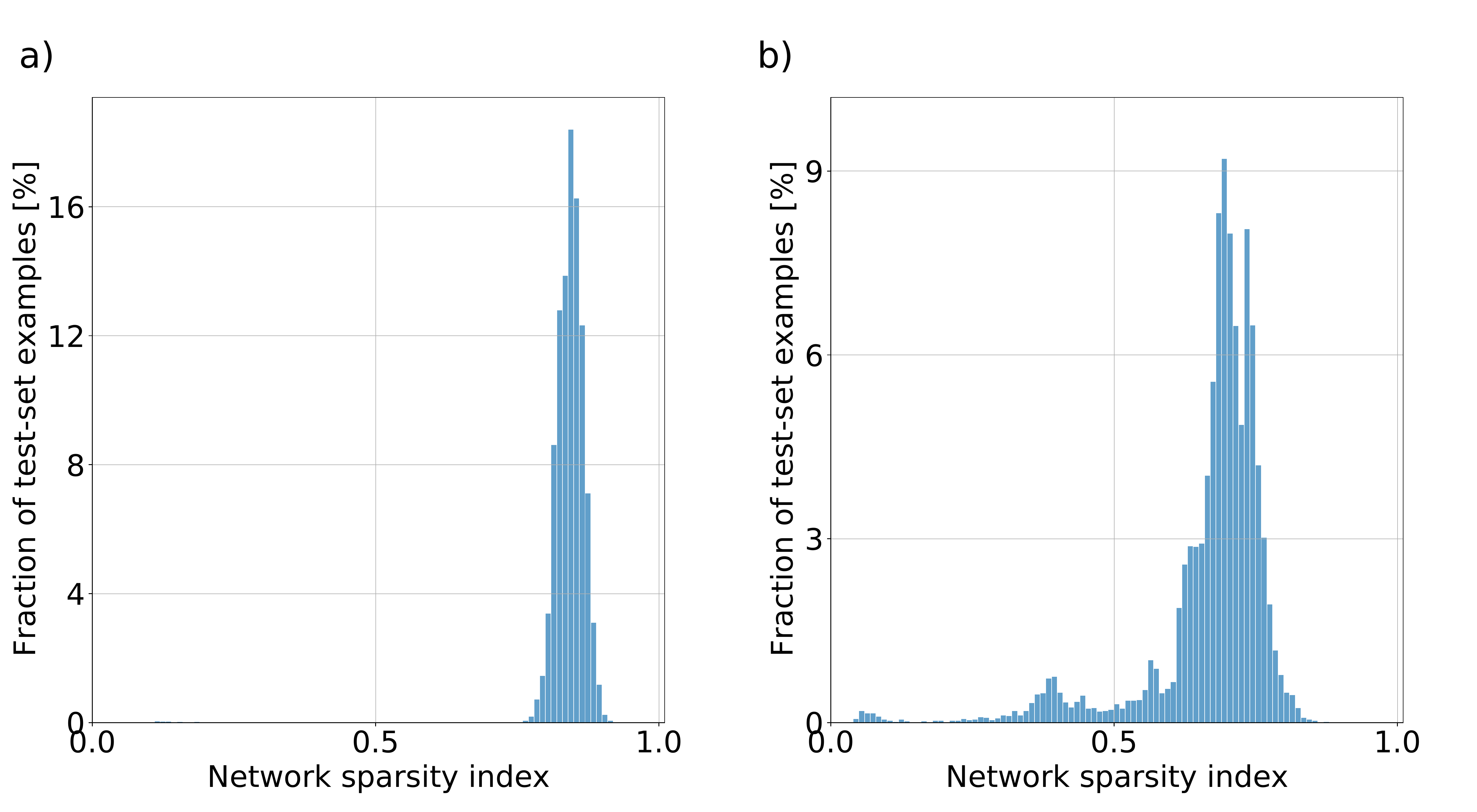}
	\centering
	\caption{Network sparsity index empirical distributions for the a)~\emph{binary} and b)~\emph{grayscale} models.}
	\label{fig:sparsity_by_model}
\end{figure}

We compute the ratio ${QN}_x$ for each example in the test-set, which we denote $QN$ for brevity. The resulting hidden layer activation sparsity empirical distribution is presented in \figurename{~\ref{fig:sparsity_by_model}}. The results suggest that this neural activity sparsity is context-based, meaning that a different subset of neurons will respond to each input signal. Only a negligible number of neurons never fire in response to any image (corresponding to $QN_x=1$ for those images), which implies that the observed sparsity is a result of the causal set neuron selection and is fundamentally different from permanently inactive neurons which can be pruned from the network. Lastly, while there seems to be a slight trade-off between classifier performance and network sparsity (as evidenced by \tablename{~\ref{tab:sparsity}}), it can be considered small given that only about 15\% of neurons are used to process each example.

\begin{table}[t!]
	\centering
	\caption{Summary of test-set-averaged classifier performance and observed network sparsity indices $QN$.}
	\label{tab:sparsity}
	\begin{tabular}{l|c|c}
		& \multicolumn{1}{c|}{F1-score} & \multicolumn{1}{c}{$QN$}  \\ \hline
		black\&white & \textbf{0.9466} & 0.0013                  \\
		binary       & 0.9386          & \textbf{0.8423}         \\
		grayscale    & 0.9238          & 0.6698                                            
	\end{tabular}
\end{table}

\subsection{Classifier Time-Performance}
\label{sec:033_results_time}

\begin{figure}[t!]
	\includegraphics[width=\figurenarrow]{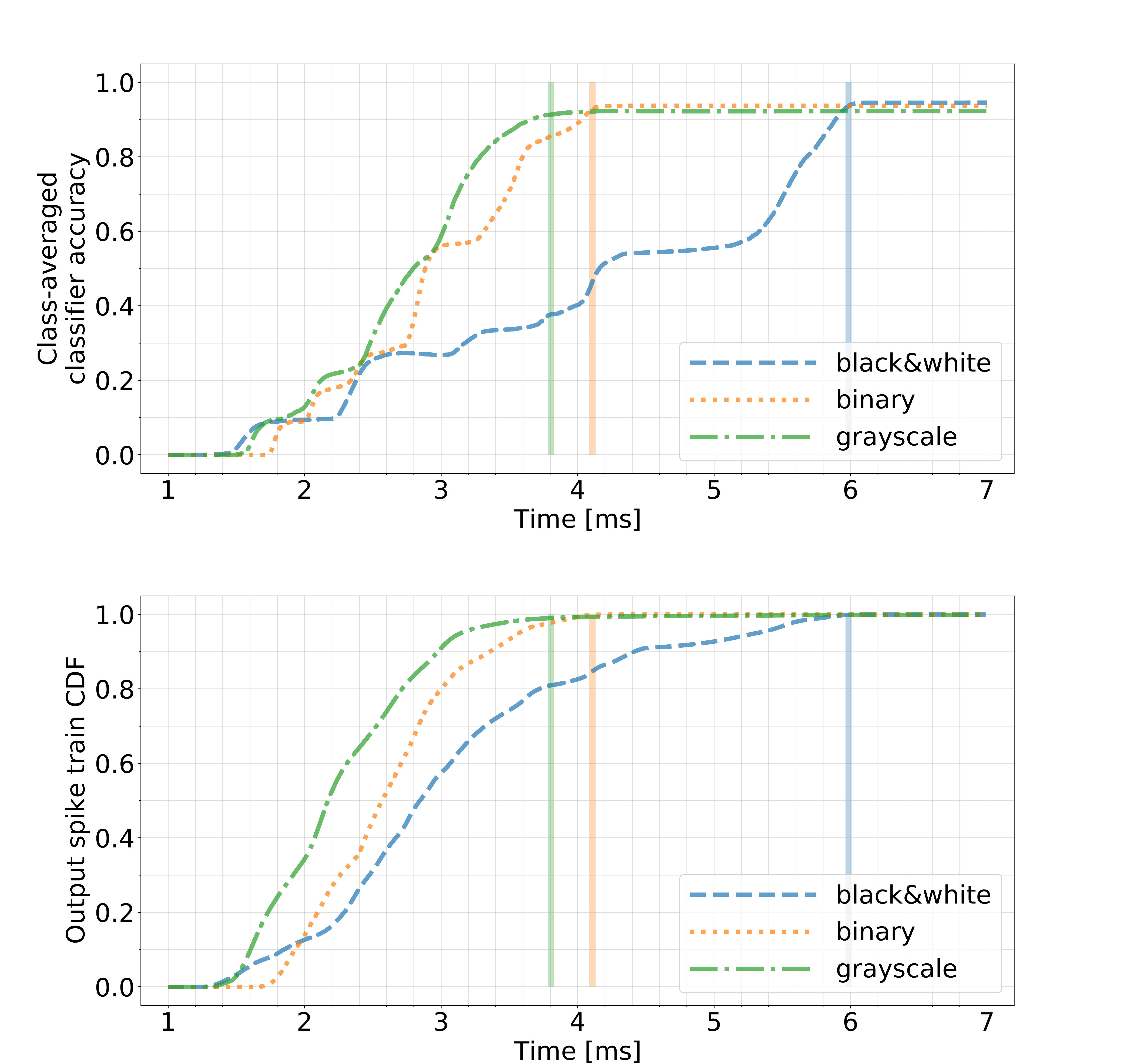}
	\centering
	\caption{Top~row: the change in class-averaged classifier accuracy over time. Bottom~row: cumulative distribution functions of output spike-times for all test-set examples, regardless of class labels. Thick vertical lines denote the time when each model reaches its steady-state performance.}
	\label{fig:time_overall}
\end{figure}

In order to investigate how the classifier performance changes as output events are observed over time, we simulate the use-case where classifier is asked to update its prediction whenever a new output event occurs by comparing with the embeddings of the training-set. This procedure provides valuable insight into prediction latency. \figurename{~\ref{fig:time_overall}} shows the class-averaged classifier accuracy for models trained with different encoding schemes. Interestingly, while the classifier performance of \emph{binary} and \emph{grayscale} models roughly correlates with the overall number of observed output spikes, the performance for the \emph{black\&white} model stops improving quite early and reaches its maximum only after observing a small number of late events. If we consider the maximum-accuracy performance of the model as its steady-state, then we find that the \emph{black\&white} model achieves steady-state about 45\%~later than the other models.

Overall, the model time-performance curves show that the quality of class label prediction increases as more output events are observed. A similar classifier accuracy vs. time study was conducted by Diehl~\emph{et~al.}~\cite{Diehl2015_fast} by describing a rate-coding artificial-to-spiking neural network conversion scheme. They report that the steepness of the time-accuracy curve depends on the network structure and input signal properties. Our results seem to give further proof to their conclusions, although we did not vary the network structure.

\section{Conclusion}
\label{sec:05_conclusions}

In this paper we presented a novel approach to supervised training of multilayer spiking siamese neural networks applied to image domain. The proposed model of a timing-sensitive spiking neural network can be trained even when some data inputs are represented by a lack of event (through implicit information). This training procedure results in models which take less time to make high-accuracy predictions and process signals using only a small subset of hidden layer neurons firing in response to the input event stream, compared to models trained with explicit event encoding. Concretely, the \emph{black\&white} model reached an \mbox{F1-score} of 0.9466 using almost all neurons to make predictions, whereas the \emph{binary} model achieved a slightly lower \mbox{F1-score} of 0.9386, however it encodes information using only 15\%~of all hidden layer neurons, and is significantly faster to reach its steady-state performance. Further investigation is warranted in order to determine whether the observed accuracy-sparsity trade-off is an artifact of our training procedure, or whether it is an inherent property of spiking neural networks. Additionally, we aim to study the effect of output layer dimensionality on these properties in subsequent research.

\bibliographystyle{IEEEbib}
\bibliography{references}

\end{document}